\documentclass[10pt,twocolumn,letterpaper]{article}

\usepackage{cvpr}
\usepackage{times}
\usepackage{epsfig}
\usepackage{graphicx}
\usepackage{amsmath}
\usepackage{amssymb}
\usepackage{booktabs}
\usepackage{caption}
\usepackage{subcaption}

\usepackage{xcolor}
\usepackage{colortbl}

\definecolor{sh_gray}{rgb}{0.84,0.84,0.84}
\definecolor{sh_gray2}{rgb}{1,0.89,0.75}
\definecolor{color3}{rgb}{0.95,0.95,0.95}
\definecolor{color4}{rgb}{0.96,0.96,0.86}
\definecolor{color5}{rgb}{0.90,0.90,0.90}

\newlength{\Oldarrayrulewidth}

\usepackage[pagebackref=true,breaklinks=true,letterpaper=true,colorlinks,linkcolor=blue,citecolor=blue,urlcolor=blue,bookmarks=false]{hyperref}

\cvprfinalcopy 

\def\cvprPaperID{} 

\ifcvprfinal\fi

\usepackage{cuted}
\usepackage{capt-of}

\begin{document}
\vspace{-1em}\title{Low Light Image Enhancement via Global and Local Context Modeling}



\author{Aditya Arora$^{1}$ \quad Muhammad Haris$^{2}$ \quad Syed Waqas Zamir$^{1}$ \quad Munawar Hayat$^{3}$ \\ 
Fahad Shahbaz Khan$^{2}$ \quad Ling Shao$^{1}$ \quad Ming-Hsuan Yang$^{4,5}$ \\
$^1$Inception Institute of AI, UAE \quad
$^2$Mohamed bin Zayed University of AI, UAE \\
$^3$Monash University, Australia \quad $^4$University of California, Merced \quad $^5$Google Research  \\
}

\maketitle

\begin{figure*}[t]
  \begin{center}\scalebox{0.95}{
    \begin{tabular}{cccc}
      \includegraphics[width=0.24\textwidth]{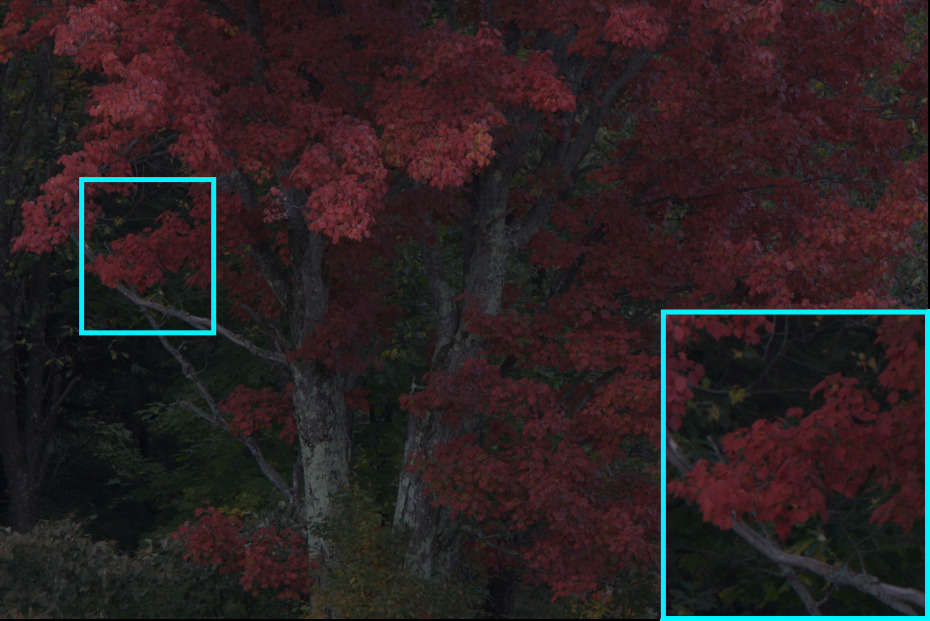}&\hspace{-4mm}
      \includegraphics[width=0.24\textwidth]{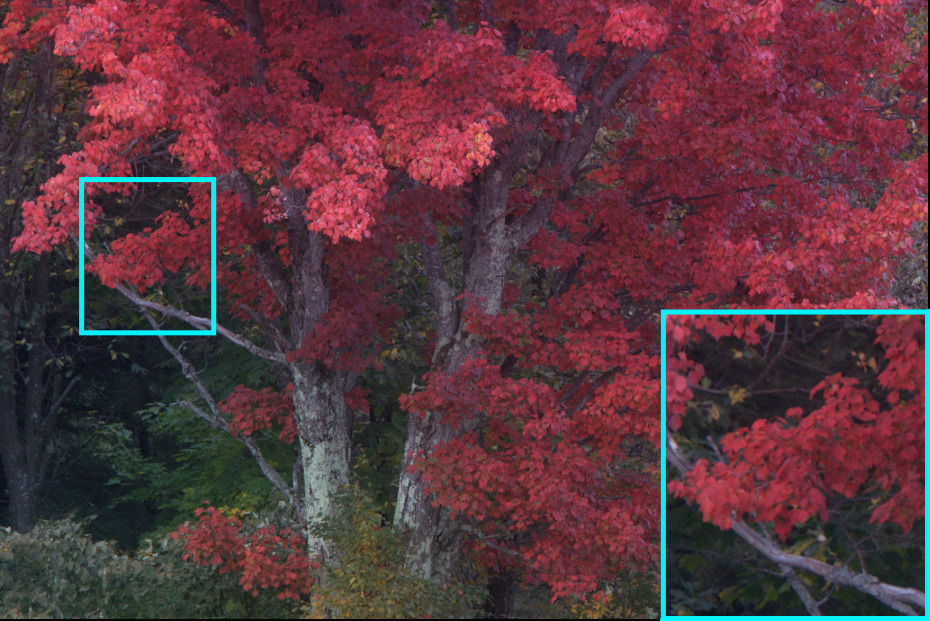}&\hspace{-4mm}
      \includegraphics[width=0.24\textwidth]{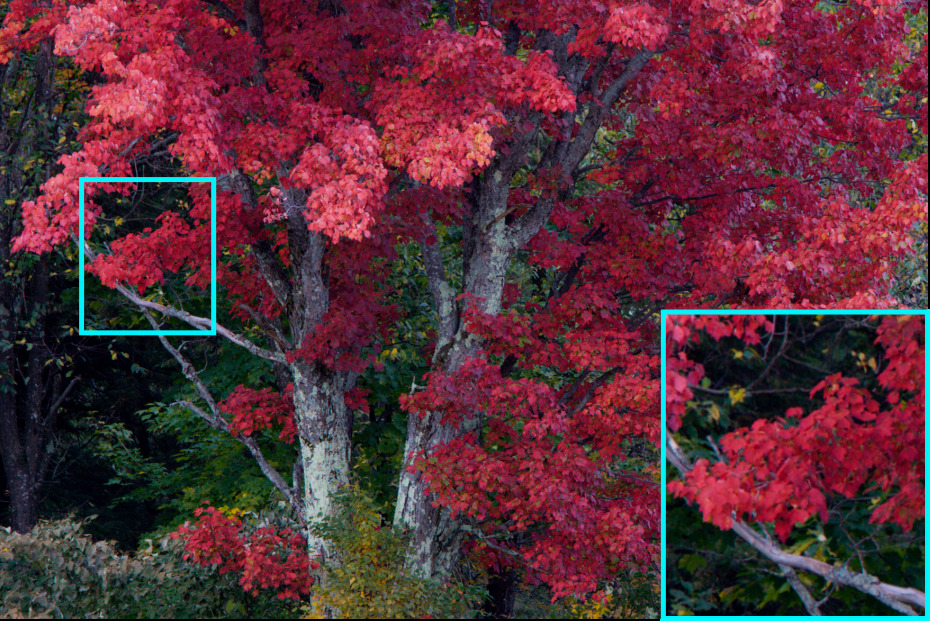}&\hspace{-4mm}
      \includegraphics[width=0.24\textwidth]{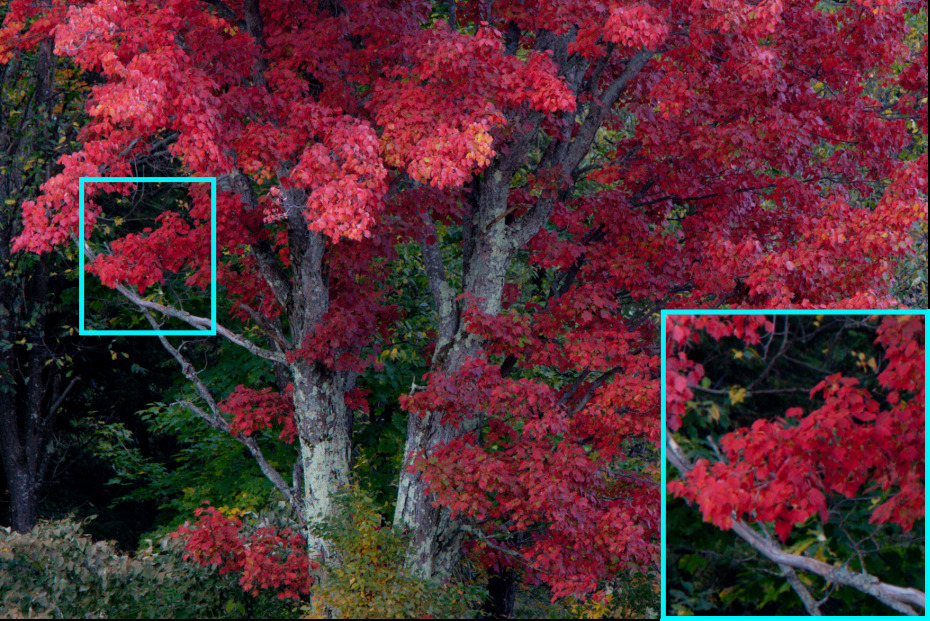}\hspace{-4mm}\\
      \includegraphics[width=0.24\textwidth,height=0.04\textwidth]{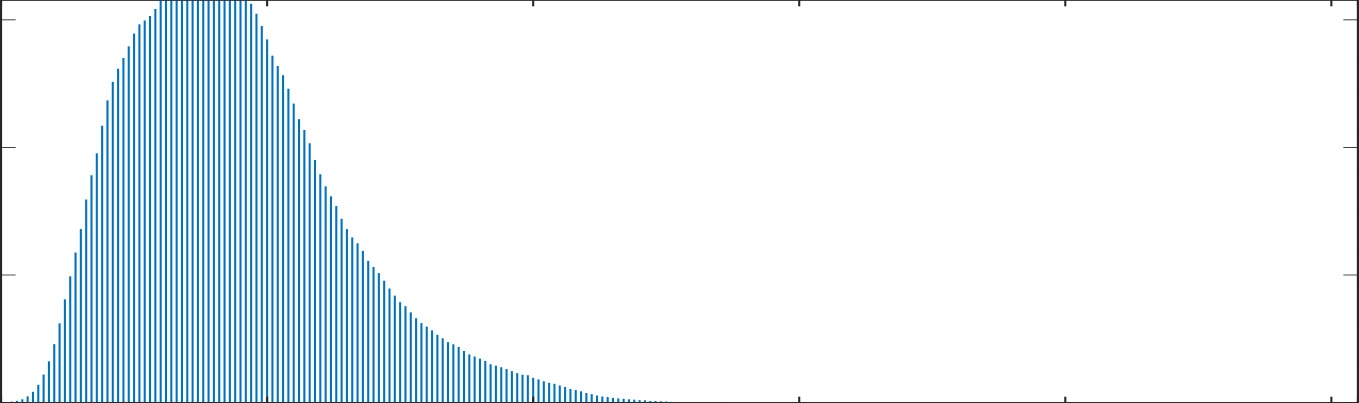}&\hspace{-4mm}
      \includegraphics[width=0.24\textwidth,height=0.04\textwidth]{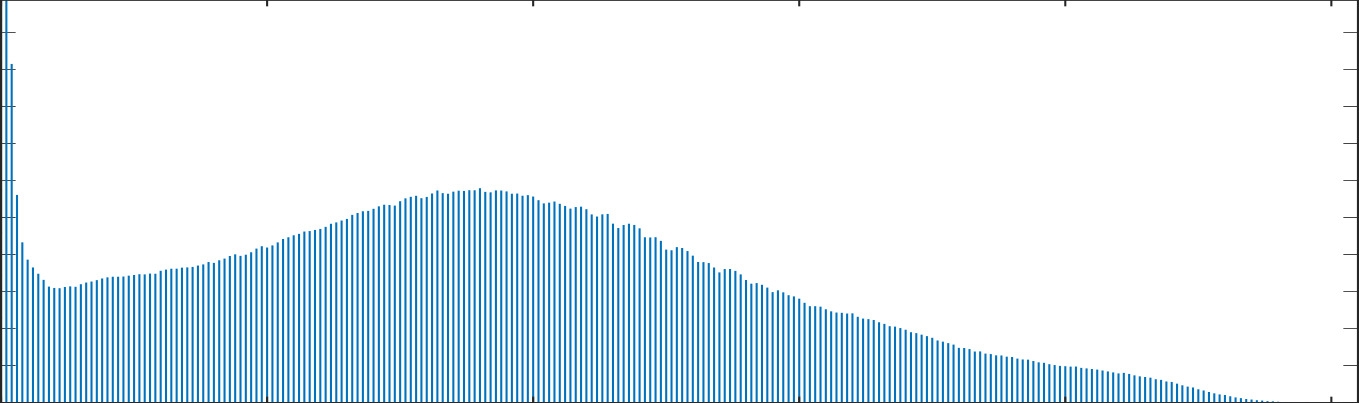}&\hspace{-4mm}
      \includegraphics[width=0.24\textwidth,height=0.04\textwidth]{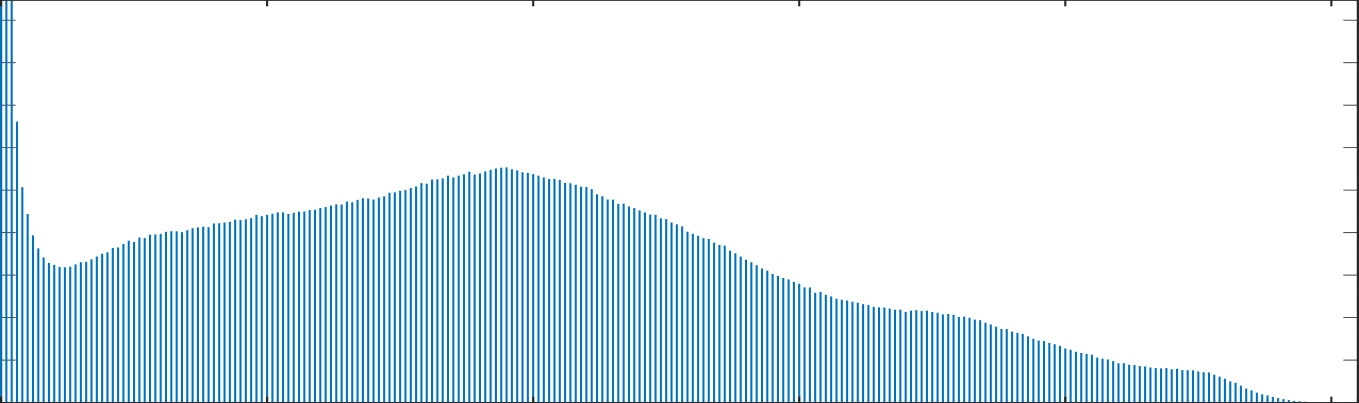}&\hspace{-4mm}
      \includegraphics[width=0.24\textwidth,height=0.04\textwidth]{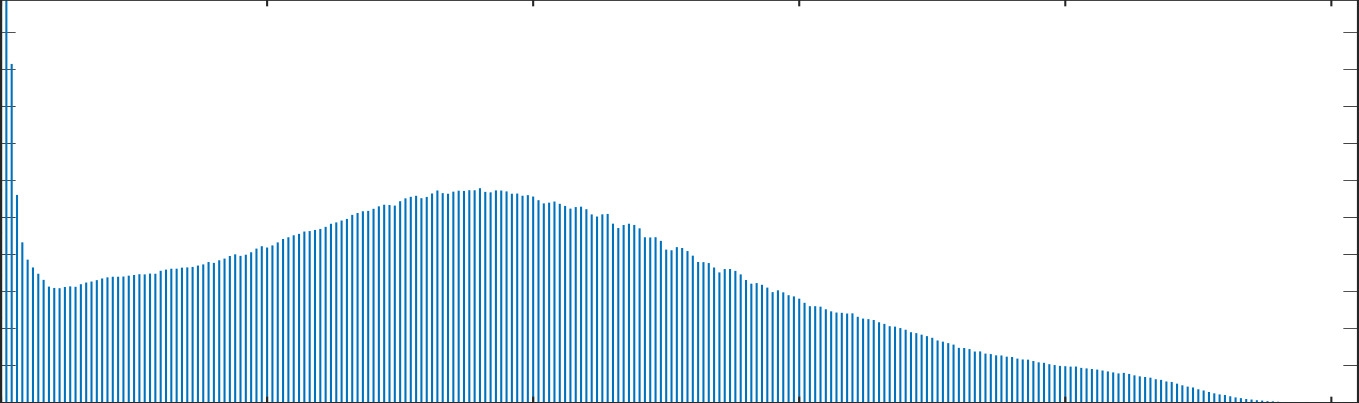}\hspace{-4mm}\\
      (a) Low-Light Input & (b) DeepUPE \cite{wang2019underexposed} &
      (c) Ours  & (d) Ground-truth \\
    \end{tabular}}
  \end{center}\vspace{-1.6em}
      \caption{\small Visual comparison of our approach with the  state-of-the-art method  DeepUPE \cite{wang2019underexposed}. 
      The proposed algorithm exhibits better local and global contrast (see histograms) with more faithful details to the ground-truth.}
      \vspace{-1.4em}
    \label{Fig:teaser}
\end{figure*}

\begin{abstract}\vspace{-1em}
Images captured under low-light conditions manifest poor visibility, lack contrast and color vividness. 
Compared to conventional approaches, deep convolutional neural networks (CNNs) perform well in enhancing images. 
However, being solely reliant on confined fixed primitives to model dependencies, existing data-driven deep models do not exploit the contexts 
at various spatial scales to address low-light image enhancement. 
These contexts can be crucial towards inferring several image enhancement tasks, \textit{e.g.}, local and global contrast, brightness and color corrections; which requires cues from both local and global spatial extent.
To this end, we introduce a context-aware deep network for low-light image enhancement. 
First, it features a global context module that models spatial correlations to find complementary cues over full spatial domain. 
Second, it introduces a dense residual block that captures local context with a relatively large receptive field.
We evaluate the proposed approach using three challenging datasets: MIT-Adobe FiveK, LoL, and SID. 
On all these datasets, our method performs favorably against the state-of-the-arts in terms of standard image fidelity metrics. 
In particular, compared to the best performing method on the MIT-Adobe FiveK dataset, our algorithm improves PSNR from 23.04 dB to 24.45 dB.
\end{abstract}

\vspace{-0.5em}
\section{Introduction}
In well-lit conditions, consumer DSLR and smartphone cameras capture reasonably good quality images.
However, in dimly-lit scenes, they often yield images that are noisy, without scene details, and manifest poor color and contrast. 
These degraded images pose challenges to various fundamental tasks in computer vision, such as semantic segmentation, object detection and tracking. 
Thus, it is of great importance to develop effective image enhancement methods to generate higher-quality images from the degraded inputs.

To capture well-exposed images in low light conditions, one may use high ISO, long exposures, and camera flash. %
However, such techniques are limited in various aspects \cite{zhang2019kindling}. 
For instance, increasing ISO can improve brightness but at the expense of noise amplification. 
Long exposure is limited to stationary scenes, and slight camera and/or object motion can potentially result in a blurry image. 
On the other hand, flashes can light up the scenes but usually introduces unwanted highlights and unbalanced illumination. 
Above all, it is likely for a user with constrained photographic devices \textit{e.g.}, typical mobile phone cameras, does not have access to these options.
Interactive methods for image enhancement allow users to adjust photos. 
They are, however, difficult and tedious for ordinary users. 
These methods require simultaneous manipulation of different parameters relating to color and contrast \cite{wang2019underexposed}. 
Semi-automatic methods alleviate this problem to an extent by reducing the number of adjustable parameters.
As these methods are sensitive to parameters and developed based on heuristic rules, 
the quality of the enhanced images is likely low \cite{chen2018deep}.  

Several algorithmic advances have been made to address this problem. 
Initial attempts \cite{cai2017joint,fu2016weighted,wang2017contrast,yuan2012automatic,zuiderveld1994contrast} mainly focus on enhancing contrast, and are limited in recovering color and local details. 
Recently, deep CNNs yield promising results by learning adjustments of color, contrast, brightness and saturation simultaneously.
Towards this end, Chen \etal~\cite{Chen2018} develop a low-light image dataset and a data-driven camera processing pipeline, and Ran \etal~\cite{ren2019low} propose a two-stream architecture through combining encoder-decoder network and a Recurrent Neural Network (RNN). 
Based on the Retinex model, Zhang \etal~\cite{zhang2019kindling} decompose an image into illumination and reflectance components, whereas the method by Wang \etal~\cite{wang2019underexposed} learn image-to-illumination mapping. 
Despite significant progress, these methods may not generate optimal images in terms of brightness, color and contrast see Figures~\ref{Fig:teaser},  \ref{Fig:qual_fivek2_2}, \ref{Fig:qual_lol}, and~\ref{Fig:qual_sid}).

A plausible explanation for this can be attributed to the reliance on local operators to model dependency across different regions of the (degraded) input. 
Local operators, typically in the form of a convolution kernel, have a fixed, local receptive field. 
It cannot potentially infer several adjustments and remove artifacts that potentially demand involvement of correlated information, termed \emph{context} hereafter, scattered at various spatial scales. 
Contextual information is known for playing a pivotal role towards improving the performance of numerous high-level computer vision tasks \cite{chen2017spatial,ding2018context,liu2018structure,yao2010modeling}. 
However, its potential is not fully harnessed for the data-driven based low-light image enhancement methods. 
We show that modeling context flexibly at various scales can provide complementary cues, which are not typically available with local receptive fields.

Global context leverages full spatial extent to search for correlated cues in the hope of enhancing a given feature. 
For example, to be able to infer global contrast for a given pixel on some foreground object, we may need to refer distant background regions. %
Further, the flexibility of querying full spatial extent and adjusting response accordingly allows effective local adjustments based on overall lighting conditions and scene settings. 
Aside from global context, information surrounding immediate vicinity of a certain spatial location can facilitate the model to better handle difficult regions.
For instance, an underexposed image may have some regions (\textit{e.g.}, dark) that are difficult to reconstruct reliably relative to others (\textit{e.g.}, bright). 
To recover color in a dimly-lit car frontal region, we may need to also encode visual cues from the well-exposed similar appearing local regions.

\vspace{0.3em}\noindent\textbf{Contributions.} We propose a context-aware network for low-light image enhancement. 
First, the proposed method exploits a \emph{global context} module for adjusting to the overall scene settings via querying full spatial extent and recalibrating accordingly for a given feature. 
Second, the proposed method introduces \emph{local context} modeling; it offers a relatively larger field-of-view with dense feature sampling. 
The local context modeling improves context diversity by complementing its global context counterpart. 
The local context modeling encodes relatively large contextual representation in a dense manner, thereby better recovering local details in difficult regions of a degraded image for reliable reconstruction.
We validate our method on three challenging image enhancement benchmarks: MIT-Adobe FiveK \cite{bychkovsky2011learning}, LoL \cite{wei2018deep}, and SID \cite{Chen2018}. 
We perform a thorough ablation to demonstrate the effectiveness of harnessing learnable context towards low-light enhancement problem. 
Experimental results demonstrate that our approach performs favorably against baselines without using contextual information.
When compared to the state-of-the-art method \cite{wang2019underexposed} on the MIT-Adobe FiveK dataset, our algorithm achieves performance gain from 23.04 dB to 24.45 dB in terms of PSNR.
Finally, the novel incorporation of context into an image enhancement network is likely to be adopted by future work and will serve as a step to develop new context-aware mechanisms for better and effective image enhancement.
\section{Related Work}
\textbf{Image Processing Methods.} An intuitive approach to enhance the visibility in a low-light image is via increasing brightness.
Albeit simple, this approach may amplify noise and generate strong color casts.
Histogram equalization based methods aim to make the image histogram uniform. 
However, histogram equalization method often produces results that are either under-enhanced or over-enhanced \cite{bertalmio2007}.
\\
\indent Instead of globally adjusting contrast, more informed approaches make use of illumination concept. The enhancement methods based on the Retinex theory \cite{land1977retinex} decompose an image into a pixel-wise product of an illumination and reflectance components, \textit{e.g.}, 
Single-scale Retinex (SSR) \cite{jobson1997properties} and Multi-scale Retinex (MSR) \cite{jobson1997multiscale}. 
However, these methods often generate unnatural outputs. 
Wang \etal \cite{wang2013naturalness} propose to jointly enhance the contrast and conserve illumination, and Guo \etal \cite{guo2016lime} recover a structured illumination map from an initial one.
These methods do not explicitly deal with noise and color distortions while enhancing images. 
In \cite{fu2016weighted}, Fu \etal propose a weighted variational model for simultaneous estimation of illumination and reflectance.
Li \etal \cite{li2018structure} extend \cite{guo2016lime} by introducing an additional term to account for extra noise. 
These two approaches \cite{fu2016weighted,guo2016lime} are able to deal with noise to some extent, but may not effectively correct color distortions.

\vspace{0.3em}
\noindent\textbf{Deep Learning Methods.} 
Numerous CNN models have been developed for low-level vision tasks, e.g., denoising \cite{Lefkimmiatis2018,Plotz2017}, demosaicking \cite{Kokkinos2018}, restoration \cite{zhang2019residual}, and dehazing \cite{cai2016dehazenet}. 
Specifically for low-light imaging, Lore \etal \cite{lore2017llnet} propose a stacked-sparse denoising autoencoder for both contrast enhancement and denoising. 
Shen \etal \cite{Shen2017} show that the multi-scale Retinex model is equivalent to a feedforward CNN with different Gaussian kernels and propose the MSR-Net to learn the mapping from dark to bright images. 
Based on the Retinex model, Wei \etal \cite{wei2018deep} develop the Retinex-Net that performs decomposition followed by illumination adjustment for image enhancement. 
Recently, generative adversarial networks (GAN) have been used for image enhancement. 
Chen \etal \cite{chen2018deep} present an unpaired learning model in a two-way GAN framework, while Ignatov \etal \cite{ignatov2018wespe} design a weakly-supervised GAN-based model. 
In addition, Deng \etal \cite{deng2018aesthetic} develop an aesthetic-driven enhancement method via adversarial learning.  

\indent With the aim of learning camera imaging pipeline, Chen \etal~\cite{Chen2018} train an encoder-decoder network. 
Ren \etal \cite{ren2019low} develop an architecture by combining an encoder-decoder and a RNN for low-light enhancement. 
Inspired by the Retinex theory, Zhang \etal \cite{zhang2019kindling} decompose an image into an illumination component for light adjustment and a reflectance component for degradation removal. 
Wang \etal \cite{wang2019underexposed} pose the recovery of the reflectance component of a degraded image as the enhanced image and propose to learn image-to-illumination mapping. 
Despite significant progress, the aforementioned approaches may not perform well in cases that require complementary cues from the full spatial extent. 
To this end, we introduce a context-aware CNN that imposes global consistency with the capability of encoding richer contextual features densely.

\vspace{0.3em}\noindent\textbf{Contextual cues in vision tasks.} Contextual cues have been widely used for vision problems, such as detection \cite{chen2017spatial,liu2018structure} and semantic segmentation \cite{ding2018context,peng2017large}.~Recently, \emph{learnable context} has been shown to be effective for numerous high-level vision tasks \cite{girdhar2017attentional,liu2018structure,peng2017large,yu2018generative}. 
Here, we show that contexts (both global and local) can be better exploited in data-driven models for low-light image enhancement.

\begin{figure*}[t!]
\begin{center}
 \includegraphics[width=0.85\linewidth]{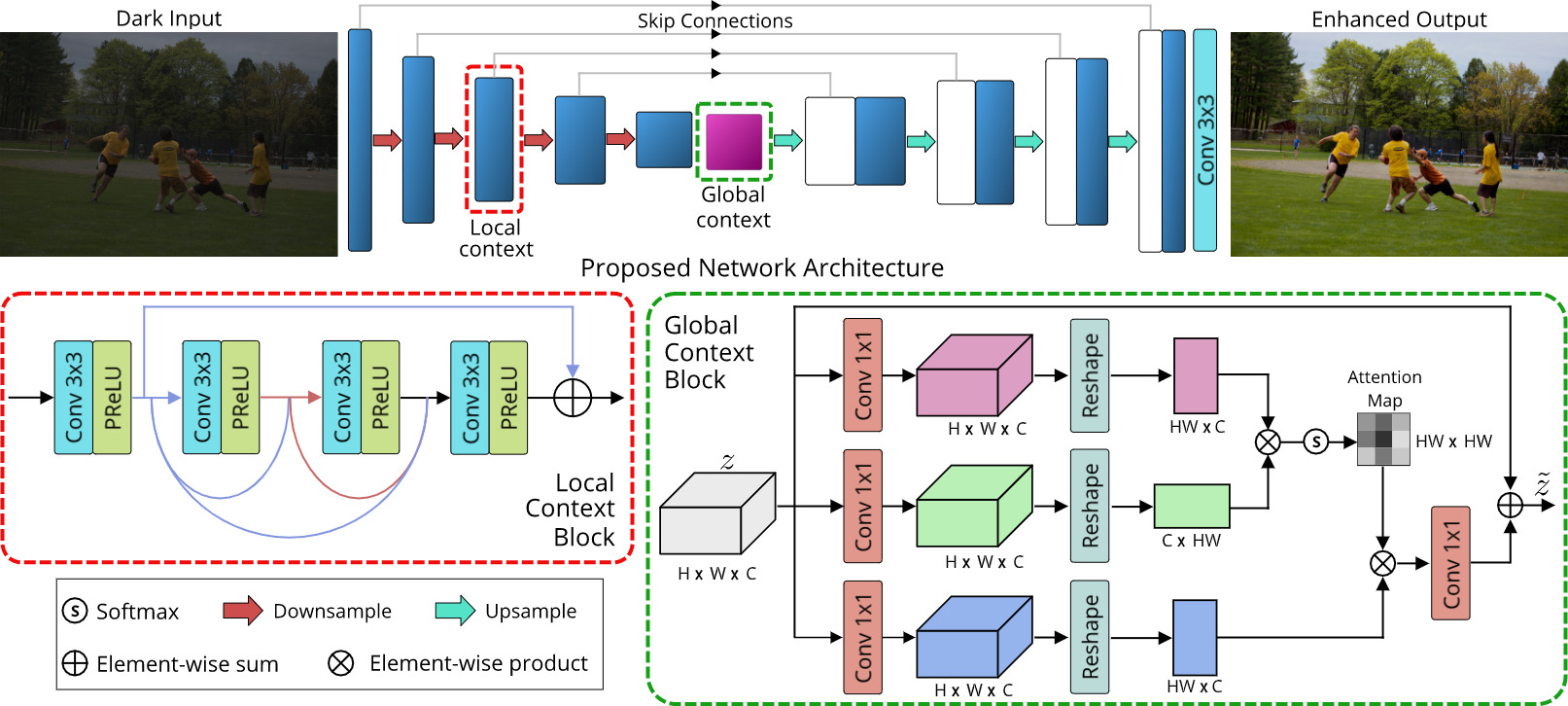}  
\end{center}\vspace{-1.2em}
    \caption{Overall network architecture for low-light image enhancement. It takes a degraded image (an underexposed photo) as input, and reproduces a  well-exposed image as output. 
    At its core, it is an encoder-decoder formulation.
    First, we introduce global context module, highlighted with dotted green box, to resolve inconsistencies requiring distant complementary cues, thereby utilizing full spatial domain. 
    Second, we introduce a dense residual block as an effective replacement to the basic block, shown with dotted red, to capture local context at various resolutions in an encoder-decoder architecture.}
    \label{Fig:overall_framework} \vspace{-1.4em}
\end{figure*}


\section{Method}

Figure~\ref{Fig:overall_framework} shows the overall architecture of our context-aware deep network.~It takes a degraded image (an underexposed photo) as an input, and reproduces a high quality, well-exposed image as output. 
Fundamentally, it is an encoder-decoder formulation built upon small convolutions (\textit{e.g.}, 3$\times$3) and skip connections. We briefly describe encoder-decoder architecture in the following section. 

\vspace{0.3em}\noindent\textbf{Encoder-Decoder Architecture.} 
Let $x$ be the short-exposure (sRGB) image captured in low-light and $\hat{x}$ be the enhanced output image. 
At the encoding stage, the input is downsampled steadily after generating features eventually bringing it to the lowest possible resolution. 
We can represent encoder latent output $z$ as:
\begin{equation}
z = e(x),
\label{Eq:encoder_side} 
\end{equation}
\noindent where $e$ is the encoding function. It can be expressed as: 
\begin{equation}
e(x) = s_{e,m}(s_{e,m-1}(s_{e,0})),
\label{Eq:encoder_side_exp} 
\end{equation}

\noindent where $s_{e,m}$ denotes the $m^{th}$ stage of the encoder. It comprises a basic block, denoted as $bb$, for computing new features followed by a maxpool operation for downsampling. 
\begin{equation}
s_{e,m} = pool_{e,m}(bb_{e,m}(x)).
\label{Eq:encoder_side_exp_res} 
\end{equation}

\noindent After computing new features $bb_{e,m}$ and before downsampling, the signal is bypassed with a skip connection, \textit{i.e.}, $skip_{e,m} = bb_{e,m}$ to the decoder side to compensate for the missing details while spatial decoding. 

When decoding, the encoder output is gradually upsampled after concatenating skip features to reconstruct the output with the same resolution as input. 
The decoder output can be written as:
\begin{equation}
\hat{x} = d(z),
\label{Eq:decoder_side}
\end{equation}

\noindent where $d$ is the decoding function, further expanded as:
\begin{equation}
d(z) = s_{d,m}(s_{d,m-1}(s_{d,0})),
\label{Eq:decoder_side_exp}
\end{equation}
\noindent where $s_{d,m}$ denotes the $m^{th}$ stage of the decoder. 
It first upsamples the input, concatenates it with the bypass (skip) output from the encoder side, and finally generates new features using the basic block:
\begin{equation}
s_{d,m} = bb_{d,m}([up_{d,m}(z), skip_{e,m}]),
\label{Eq:decoder_side_exp_res}
\end{equation}
where $[\cdot]$ is a concatenation operator. 
We use the encoder-decoder based architecture, including skip connections, as our baseline framework for low-light image enhancement.

\subsection{Proposed Context-aware Network}
Encoder-decoder architectures \cite{long2015,newell2016stacked,ronneberger2015u} extract deep features at various resolutions to encode the input $x$ into a latent representation $z$, before performing spatial decoding to generate the output $\hat{x}$. 
When applied to low-light image enhancement problems the (expected) output is a well-exposed image. 
Even after successive downsampling, these models rely on convolutional operators to improve the view of the input, to generate $z$. 
As such, these models have limited capacity in exploiting the full spatial domain.
To this end, we introduce the global context (GC) model.
Furthermore, we show that the basic block in these network formulations, responsible for extracting new features, is limited in terms of receptive field size and rigorous input sampling. 
Notably, this block is repeated multiple times (in each stage of the encoder as well as the decoder).
Therefore, addressing this issue is important for the problem studied in this work. 
For instance, it can facilitate enhancing the color and contrast of local regions in the input image. 
To accomplish this goal, we propose a local context (LC) model.

\vspace{0.3em}
\noindent\textbf{Global Context (GC).}
Deep CNNs for low-light image enhancement are mainly built upon convolutional operators \cite{Chen2018,wang2019underexposed,zhang2019kindling}. 
Being local in nature and fixed-shaped, convolutional operators are limited in modeling dependencies across features. 
However, numerous vision problems entail involvement of non-local features as complementary cues for accurate processing.  
For instance, inferring global contrast of a foreground pixel requires considering certain distant background regions.
Notably, these operations cannot be achieved with a limited view of the input image. 

Although stacking several convolutional layers may alleviate this issue and capture greater context, it also brings a number of different challenges. 
First, it makes the model unnecessarily deep and large, thereby requiring higher computational load and bigger memory in addition to the increased over-fitting risks. 
Second, information of features that are distant from a location need to be passed through several layers before affecting the location for both forward propagation and backward propagation, thereby making the optimization more complex \cite{chen20182}. 

The integration of contextual information, within deep CNNs, has been investigated in different vision tasks \cite{hu2018squeeze,hu2018gather,zhao2018psanet}. Most of these approaches, such as \cite{hu2018squeeze} summarize the input information across spatial dimensions and rescale channel features to acquire global context.
While simple and efficient, they assume the same mask for each spatial position. 
To this end, we propose to capture global context via non-local blocks (see Figure~\ref{Fig:overall_framework}) which are designed to model long-range dependencies in video/image classification \cite{wang2018non}. To our knowledge, we are the first to propose the utilization of non-local blocks to capture global context for low-light image enhancement.~A non-local block strengthens a given feature by aggregating features from the full spatial extent via a pairwise affinity measure. 

We denote $z=\left\{z_{i}\right\}_{i=1}^{N_{p}}$ as the feature map of an input tensor where $z$ is the latent representation (Eq.~\ref{Eq:encoder_side}), and $N_{p}$ is the total number of positions in a feature map, \textit{i.e.}, the full spatial extent of a feature map ($N_{p}=H \times W$). 
With $z$ and $\tilde{z}$ (of the same dimension) representing the input and output of the non-local block,  
we have:
\begin{equation}
\tilde{z}_{i} = z_{i} + W_{u}\sum_{j=1}^{N_{p}}\frac{f(z_{i},z_{j})}{C(z)}(W_{v}.z_{j}),
\label{eq:non_local_block}
\end{equation}
where $i$ corresponds to the index of a query position and $j$ iterates through all possible positions.
In Eq.~\ref{eq:non_local_block}, $f(z_{i},z_{j})$ finds the affinity between position $i$ and $j$, and has a normalization factor $C(z)$. 
In addition, $W_{u}$ and $W_{v}$ denote linear projection matrices (implemented as 1x1 conv.). 
For notation brevity, we denote $d_{ij}=\frac{f(z_{i},z_{j})}{C(z)}$ as normalized pairwise affinity between position $i$ and $j$. 
The difference lies in how pairwise affinity $d_{ij}$ is computed.  
In our global context module, we utilize the embedded Gaussian to instantiate non-local block. Next, we describe our dense residual block for modeling local context. 

\vspace{0.3em}
\noindent\textbf{Local Context (LC).}
As discussed earlier, the basic block, designated with the generation of new features, in an encoder-decoder based architecture has limited representational capability due to small receptive-fields along both resolution and scale axis w.r.t the input.
Typically, a basic block comprises of two stacked convolutions of size 3$\times$3 to generate features. 
Let $f$ be the input feature maps, we can describe a basic block by:
\begin{equation}
bb(f) = \sigma_2(W_{2}\sigma_1(W_{1}f)),
\label{Eq:bb_expression}
\end{equation}
\noindent where $W_{1}$ and $W_{2}$ are the weight matrices for convolutions and $\sigma(\cdot)$ is the PReLU activation. 

A straightforward approach to increase the receptive-field size is to downsample the input (say $\times$2) and then use convolutions (3$\times$3) followed by upsampling operation. 
This scheme, however, causes loss of spatial information (due to downsampling and upsampling) that is important for image restoration. 
To this end, we introduce a dense residual block (DRB) to model local context (see Figure~\ref{Fig:overall_framework}) by allowing faster growth of receptive field in a dense fashion. 
It consists of 3 cascaded convolutions of kernel size 3$\times$3. 
The output of each convolution is concatenated with the input features and all the outputs from previous convolutions. 
The concatenated features are then fed to the next convolution. 
Finally, we employ a skip connection from the input to the output of this block.
Formally, we formulate each convolution layer output $y_{l}$ in a DRB as:
\begin{equation}
y_{l} = M_{k,l}([y_{l-1}, y_{l-2}, \ldots,y_{0}]),
\label{Eq:drb_layerwise}
\end{equation}
\noindent where $M_{k,l}$ is the convolutional operator with kernel size of $k$ at layer position $l$, 
and $[ \cdot ]$ is a concatenation operator that combines outputs from all preceding layers. 
The output of a DRB is comptued by cascading convolutional layers $y_{l}$:
\begin{equation}
DRB(f) = y_{l}(y_{l-1}(y_{0})).
\label{Eq:drb_expression}
\end{equation}
Note that, the DRB output (Eq.~\ref{Eq:drb_expression}) captures larger field-of-view and manifests denser features via dense connections (Eq.~\ref{Eq:drb_layerwise}) than the $bb$ output (Eq.~\ref{Eq:bb_expression}). 
Furthermore, this operation is efficiently achieved by reusing features without incurring much computational overhead.


\begin{table}[t]
\begin{center}
\caption{\small 
Ablation study. Both global and local context modeling contribute to the overall improvement.
Our method achieves an absolute gain of 1.96 dB over the baseline. 
In addition, it achieves an absolute gain of 1.65 dB over either module.}
\label{tab:ablation_results}
\vspace{-0.1cm}

\setlength{\tabcolsep}{11pt}
\scalebox{0.72}{
\begin{tabular}{lcccc}
\hline
Add-on    &  \multicolumn{3}{c}{} & Ours \\
\hline
\textit{No-add (Baseline)}        &\checkmark          &\checkmark     &\checkmark      &\checkmark\\
\textit{Global Context}       &{}         &\checkmark     &{} &\checkmark\\
\textit{Local Context}            &{}          &{}          &\checkmark &\checkmark\\
\hline
PSNR    &22.49        &22.69   &22.80    &\textbf{24.45}\\
\hline
\end{tabular}
}
\end{center}\vspace{-0.5cm}
\end{table}
\section{Experiments and Analysis}
In this section, we first describe experimental settings and then we present ablation studies and main results.

\vspace{0.4em}
\noindent \textbf{Datasets.} (1) \textbf{ MIT-Adobe FiveK} \cite{bychkovsky2011learning} contains 5000 images captured with DSLR cameras. 
All photographs feature a broad range of scenes, subjects, and lighting conditions. 
Five photography students (A/B/C/D/E) manually adjust photos to generate expert-retouched images. 
Similar to \cite{hu2018exposure,park2018distort,wang2019underexposed}, we use only the output of Expert C as the ground-truth. 
The first 4500 images are used for training and the last 500 for testing.
(2) \textbf{LoL} \cite{wei2018deep} contains 500 real camera image pairs, out of which 485 images are for training and 15 for testing. Each pair consists of a low-light source image and its corresponding well-exposed ground-truth. 
(3) \textbf{See-in-the-dark (SID)} \cite{Chen2018} dataset is captured in extreme low-light with two different cameras: a) Sony $\alpha$ 7S II with Bayer filter array, and b) Fujifilm X-T2 with X-Trans array. %
We use the Sony subset which contains 424 images for training and 88 for testing. 

\vspace{0.4em}
\noindent \textbf{Evaluated Methods.} We evaluate our method by comparing it with low-light image enhancement methods. 
The LIME method \cite{guo2016lime} estimates illumination map by attending to each pixel in color channels and exploits the structure of illumination map.
The KinD scheme \cite{zhang2019kindling} decomposes image into illumination and reflectance components via two separate networks. 
The DPE method \cite{chen2018deep} uses a two-way GAN for learning the mapping from dark to bright images. 
The DeepUPE model \cite{wang2019underexposed} learns a image-to-illumination transformation function by applying constraints on the recovered illumination map.

\begin{table*}[t]
\begin{center}
\caption{\small Quantitative comparisons of the evaluated methods on LoL dataset \cite{wei2018deep}. 
The proposed algorithm performs favorably against the state-of-the-art method with an absolute performance gain of 2.14 dB (PSNR). Bold indicates the best results.}\vspace{-0.5em}
\label{tab:quan_lol}
\setlength{\tabcolsep}{3.7pt}
\scalebox{0.72}{
\begin{tabular}{l c c c c c c c c c c c c c}
\toprule
  Method & BIMEF \cite{ying2017bio} & CRM \cite{ying2017new} & Dong \cite{dong2011fast} & LIME \cite{guo2016lime}  & MF \cite{fu2016weighted} & RRM \cite{liu2018structure} & SRIE \cite{fu2016weighted} & Retinex-Net \cite{wei2018deep} & MSR \cite{jobson1997multiscale} & NPE \cite{wang2013naturalness} & GLAD \cite{wang2018gladnet} & KinD \cite{zhang2019kindling}  & Ours \\
 \midrule
PSNR & 13.86 & 17.20 & 16.72 & 16.76 & 18.79 & 13.88 & 11.86 & 16.77 & 13.17 & 16.97 & 19.72 & 20.87 & \textbf{23.01}\\
SSIM & 0.58 & 0.64 & 0.58 & 0.56 & 0.64 & 0.66 & 0.50 & 0.56 & 0.48 & 0.59 & 0.70 & 0.80 & \textbf{0.89}\\
\bottomrule
\end{tabular}}
\end{center}\vspace{-1.8em}
\end{table*}

\begin{figure*}[t]
\centering
 \includegraphics[width=\linewidth]{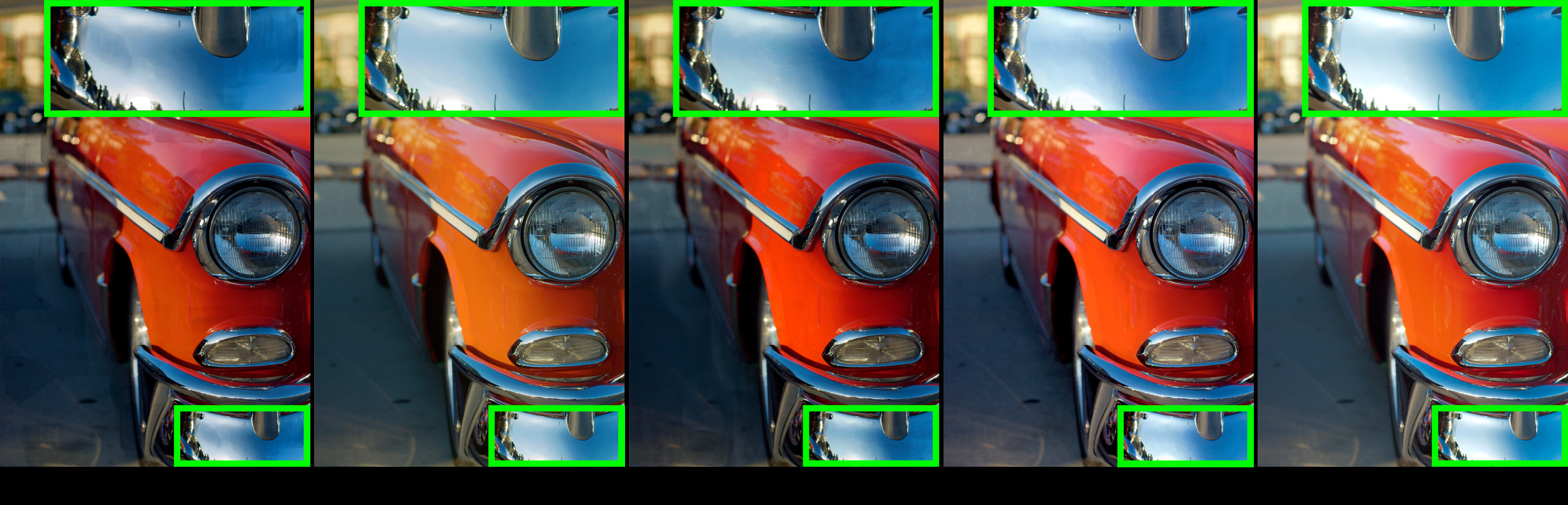} \vspace{-1.2em}
    \caption{\small Illustration of how the proposed context-aware framework.
    From left, the first image is the baseline output, revealing artifacts on the bumper and around the lower left side of the car. 
    Second image: with the inclusion of global context modeling these artifacts are removed. 
    However, it cannot faithfully recover colors at different locations \eg, behind the two lights. 
    Third image: encoding local context helps decoding of colors (closer to the target) in the spatially homogeneous regions at various car locations. 
    However, being local in nature, it does not perform adjustments at locations requiring access to global scene settings and generates artifacts.    
    Fourth image: our approach based on both global and local contexts is able to recover vivid colors at different car locations without artifacts.} 
    \vspace{-2mm} 
    \label{Fig:contributions_motivation}
\end{figure*}

\vspace{0.4em}
\noindent \textbf{Training Details.} We implement the proposed network using PyTorch with a single Titan-V100 GPU. 
The entire network is optimized using the Adam optimizer with an initial learning rate of $10^{-4}$. 
For data augmentation, we extract random crops of size $512 \times 512$ followed by random flipping and rotation.
The initial learning rate is decayed by a factor of 2 after every 128K iterations, and we train for a total number of 640K iterations.
We employ the $L_1$ loss for network optimization. 
The source code and trained models will be released publicly.

\begin{table}[t]
\begin{center}
\caption{\small Quantitative comparisons of evaluated methods on the MIT-Adobe FiveK dataset \cite{bychkovsky2011learning}. } \vspace{-0.5em}
\label{tab:quan_fivek}
\setlength{\tabcolsep}{3.0pt}
\scalebox{0.72}{
\begin{tabular}{l c c c c c c}
\toprule
  Method & HDRNet \cite{Gharbi2017} & W-Box \cite{hu2018exposure} & DR \cite{park2018distort} & DPE \cite{chen2018deep}  & DeepUPE \cite{wang2019underexposed}  & Ours \\
 \midrule
PSNR & 21.96 & 18.57 & 20.97 & 22.15 & 23.04 & \textbf{24.45}\\
SSIM & 0.866 & 0.701 & 0.841 & 0.850 & 0.893 & \textbf{0.929}\\
\bottomrule
\end{tabular}}
\end{center}\vspace{-2em}
\end{table}
\vspace{-0em}

\begin{table}[t]
\centering
\caption{\small Quantitative results on the SID dataset.
Proposed approach shows improved results compared to the method of \cite{Chen2018}.
} 
\label{tab:quan_sid}\vspace{-0.5em}
\setlength{\tabcolsep}{3.0pt}
\scalebox{0.72}{
\begin{tabular}{l c c c c c}
\toprule 
Method & Chen \etal. & Maharjan \etal. & Zamir \etal. & Karadeniz \etal. & Ours \\
 &  \cite{Chen2018} & \cite{maharjan2019improving} & \cite{zamir2019learning} & \cite{karadeniz2020burst} &  \\
\midrule 
PSNR & 28.96 & 29.17 & 28.84 & 29.29 & \textbf{29.70} \\
SSIM & 0.896 & 0.886 & 0.876 & 0.882 & \textbf{0.902} \\
\bottomrule 
\end{tabular}}\vspace{-1.3em}

\end{table}

\begin{figure*}[t]
  \begin{center}
  \scalebox{0.9}{
  \small
    \begin{tabular}{ccc}\hspace{-2mm}
      \includegraphics[width=0.325\textwidth]{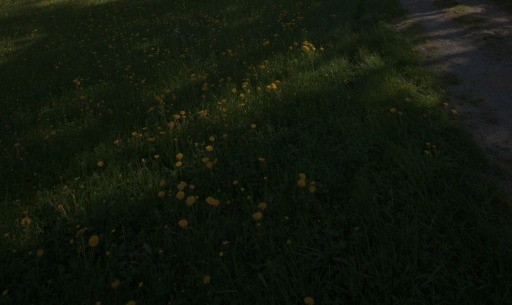}&
      \includegraphics[width=0.325\textwidth]{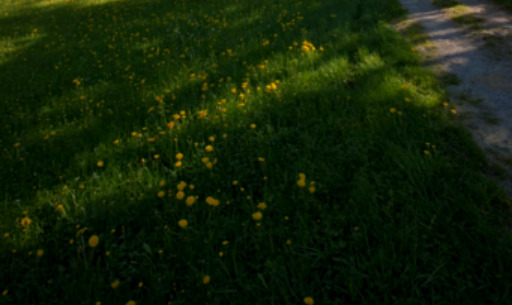}&
      \includegraphics[width=0.325\textwidth]{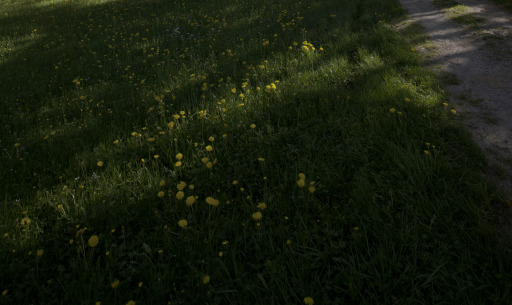}
      \\

      (a) Low-Light Input & (b) DPE \cite{chen2018deep} & (c) HDRNet \cite{Gharbi2017} \\ 
    \hspace{-2mm}
      \includegraphics[width=0.325\textwidth]{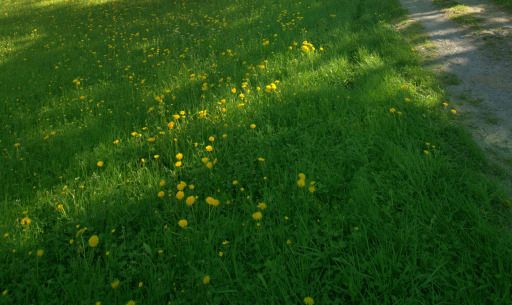}&
      \includegraphics[width=0.325\textwidth]{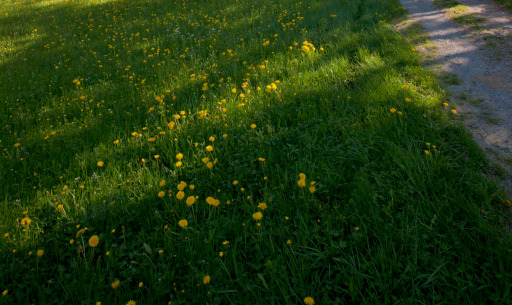}&
      \includegraphics[width=0.325\textwidth]{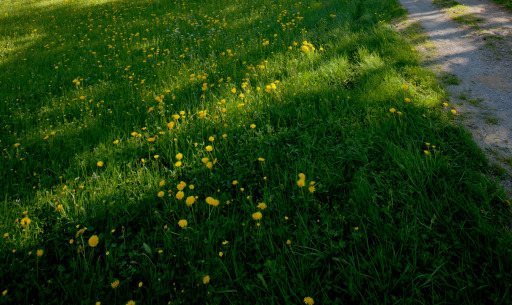}
      \\

      (d) DeepUPE \cite{wang2019underexposed} & (e) Ours & (f) Ground-Truth 
    \end{tabular}}
  \end{center}
  \vspace{-1.2em}
\end{figure*}

\begin{figure*}[t]
  \begin{center}
  \scalebox{0.9}{
  \small
    \begin{tabular}{ccc}\hspace{-2mm}
      \includegraphics[width=0.325\textwidth]{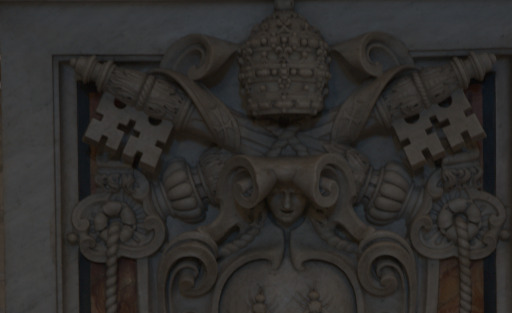}&
      \includegraphics[width=0.325\textwidth]{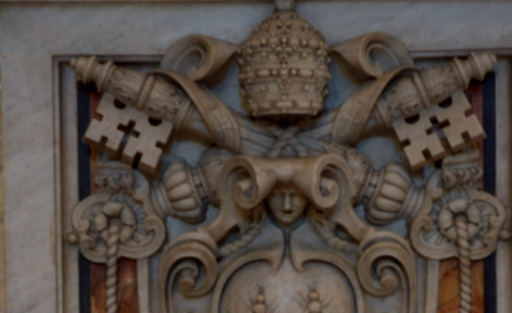}&
      \includegraphics[width=0.325\textwidth]{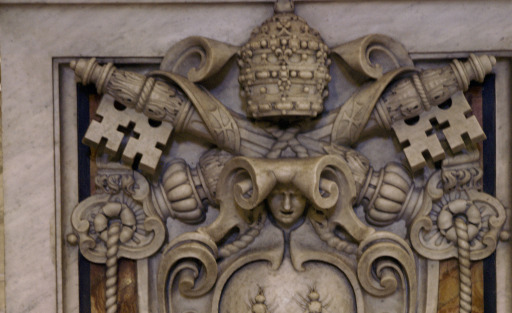}
      \\

      (a) Low-Light Input & (b) DPE \cite{chen2018deep} & (c) HDRNet \cite{Gharbi2017} \\ 
    \hspace{-2mm}
      \includegraphics[width=0.325\textwidth]{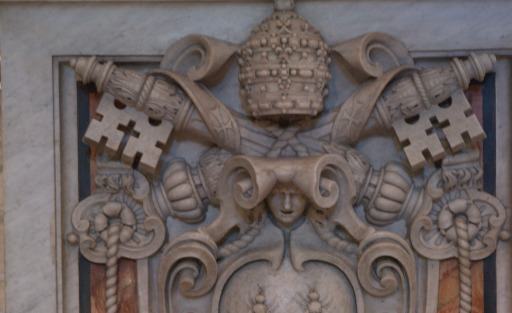}&
      \includegraphics[width=0.325\textwidth]{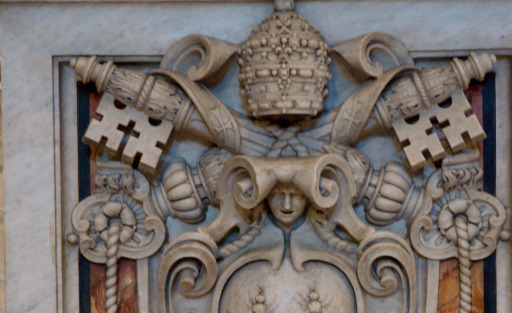}&
      \includegraphics[width=0.325\textwidth]{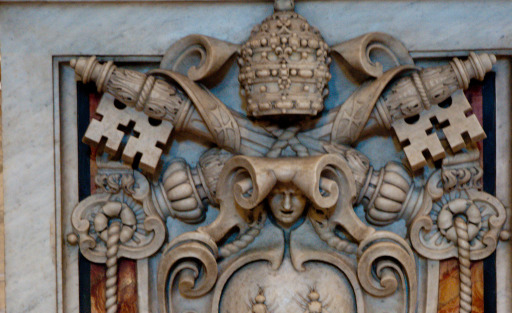}
      \\

      (d) DeepUPE \cite{wang2019underexposed} & (e) Ours & (f) Ground-Truth 
    \end{tabular}}
  \end{center}
  \vspace{-1.4em}
    \caption{\small Qualitative example from MIT-Adobe Fivek~\cite{bychkovsky2011learning}. The result produced by our method is vivid, and have better local and global contrast than other algorithms.  }
    \label{Fig:qual_fivek2_2}
    \vspace{-1.4em}
\end{figure*}

\subsection{Ablation Study}
We validate the performance impact of integrating the proposed global context modeling and the local context modeling  into the non-contextual (baseline) framework. 
Ablation experiments are performed on the MIT-Adobe FiveK \cite{bychkovsky2011learning} dataset.
Table~\ref{tab:ablation_results} shows the results from ablation studies. 
The baseline architecture (based on U-Net) achieves PSNR of 22.49~dB. 
The introduction of global context modeling improves the performance from 22.49~dB to 22.69~dB. 
We observe almost similar improvement (0.31~dB) after introducing local context modeling into baseline. 
The proposed algorithm, with both global and local context modeling, achieves an absolute performance gain of 1.96~dB in terms of PSNR in comparison to baseline. 
Note that the overall performance gain is substantially higher (by 1.65~dB) than that of using each module, individually. 
It indicates that both modules make complementary contributions for image enhancement.

In addition to the quantitative ablation results, we show enhanced images using these models in Figure~\ref{Fig:contributions_motivation}. 
The enhanced images by the baseline method contain artifacts: for instance, on the bumper and around the lower left side of the car. 
When we introduce the global context modeling in the baseline, the enhanced images no longer have artifacts; however, the color appearance do not match faithfully to the well-exposed reference images (for example, see the red color of the car). 
Encoding local context helps reproducing colors in the spatially homogeneous regions around various car locations. 
However, being local in nature, it is not effective for doing adjustments requiring global view, thereby generating some artifacts. 
Our framework, based on both GC and LC models, removes artifacts and recovers vivid colors at two different car locations, generating an enhanced image perceptually-faithful to the ground-truth.


\subsection{Quantitative Evaluation}
First, we compare the performance of our method with several other algorithms on the LoL dataset. The results are shown in Table~\ref{tab:quan_lol}. The proposed method achieves 2.14 dB improvement over the recent best method KinD \cite{zhang2019kindling}.
Next, we evaluate our method against five approaches on the MIT-Adobe FiveK dataset. 
Table~\ref{tab:quan_fivek} shows that our method performs well against all competing algorithms. Compared to the recent best method \cite{wang2019underexposed}, our method achieves 1.41~dB PSNR gain).

Finally, we evaluate the proposed method on the RAW data from the SID dataset. 
It is a more challenging case as the input to the network is now a RAW image and the output is an enhanced sRGB image. It implies that the network has to learn the complete camera imaging pipeline (which applies a series of complex operations on RAW data in order to generate sRGB images).
We use the Sony subset from the SID dataset~\cite{Chen2018} for evaluation. 
Table~\ref{tab:quan_sid} shows a comparison of results obtained with our method and with those of the state-the-art \cite{Chen2018,maharjan2019improving,karadeniz2020burst,zamir2019learning}. 
Our method achieves 0.41 dB improvement in PSNR over \cite{karadeniz2020burst}.

\begin{figure*}[t]
  \begin{center}
    \begin{tabular}{cccccc}\hspace{-3mm}
      \includegraphics[width=0.16\textwidth]{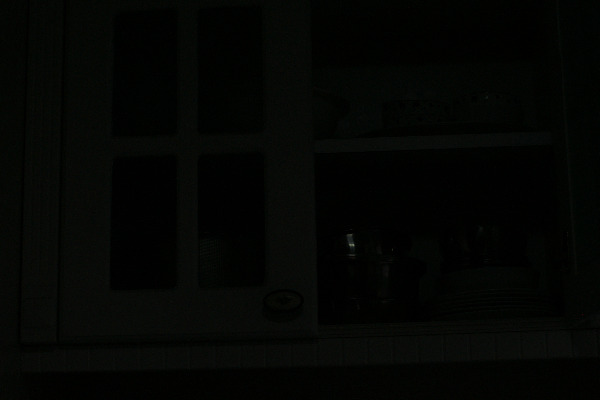}&\hspace{-4.0mm}
      \includegraphics[width=0.16\textwidth]{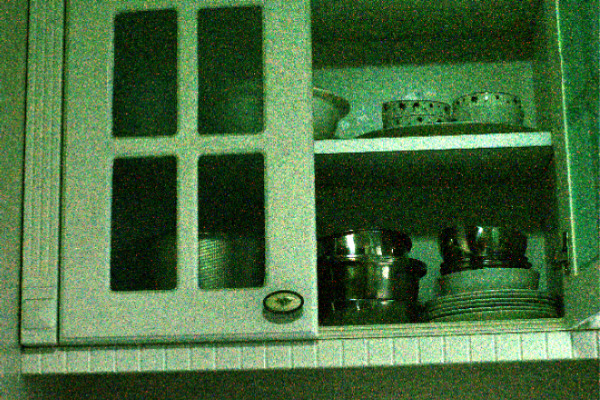}&\hspace{-4.0mm}
      \includegraphics[width=0.16\textwidth]{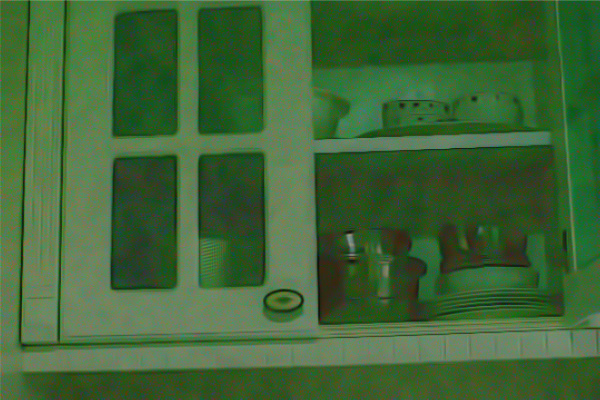}&\hspace{-4.0mm}
     \includegraphics[width=0.16\textwidth]{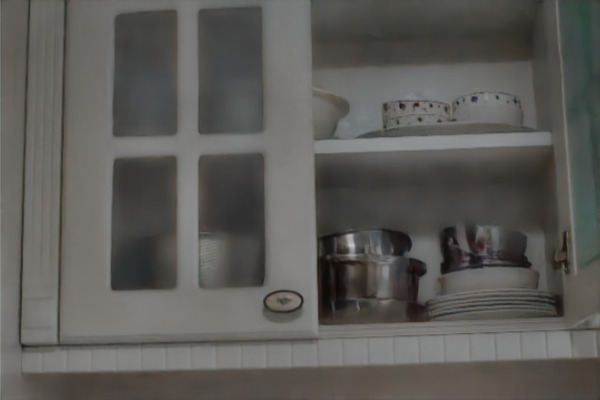}&\hspace{-4.0mm}
      \includegraphics[width=0.16\textwidth]{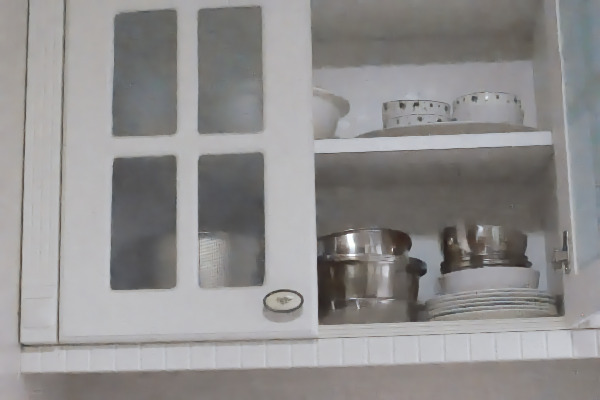}&\hspace{-4.0mm}
      \includegraphics[width=0.16\textwidth]{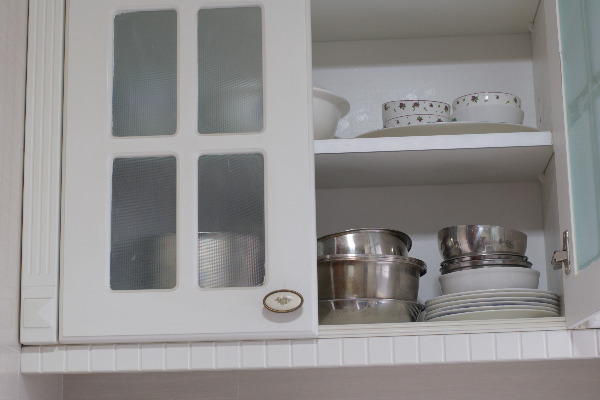}\\
      \hspace{-3mm}(a) Input & \hspace{-3mm} (b) LIME \cite{guo2016lime} & \hspace{-3mm}(c) RetinexNet \cite{wei2018deep} & \hspace{-3mm} (d) KinD \cite{zhang2019kindling} & \hspace{-3mm} (e) Ours & \hspace{-3mm}(f) Ground-Truth
    \end{tabular}
  \end{center}\vspace{-2em}
    \caption{\small Qualitative comparison of our method with state-of-the-art approaches on the LoL dataset. 
      Our method reconstructs image with distinct contrast and natural brightness. The object-level details in our results are more apparent than those of the other approaches.}
      \vspace{-0em}
    \label{Fig:qual_lol}\vspace{-0.6em}
\end{figure*}

\begin{figure*}[!t]
  \begin{center}
    \begin{tabular}{cccc}\hspace{-6mm}
      \includegraphics[width=0.242\textwidth]{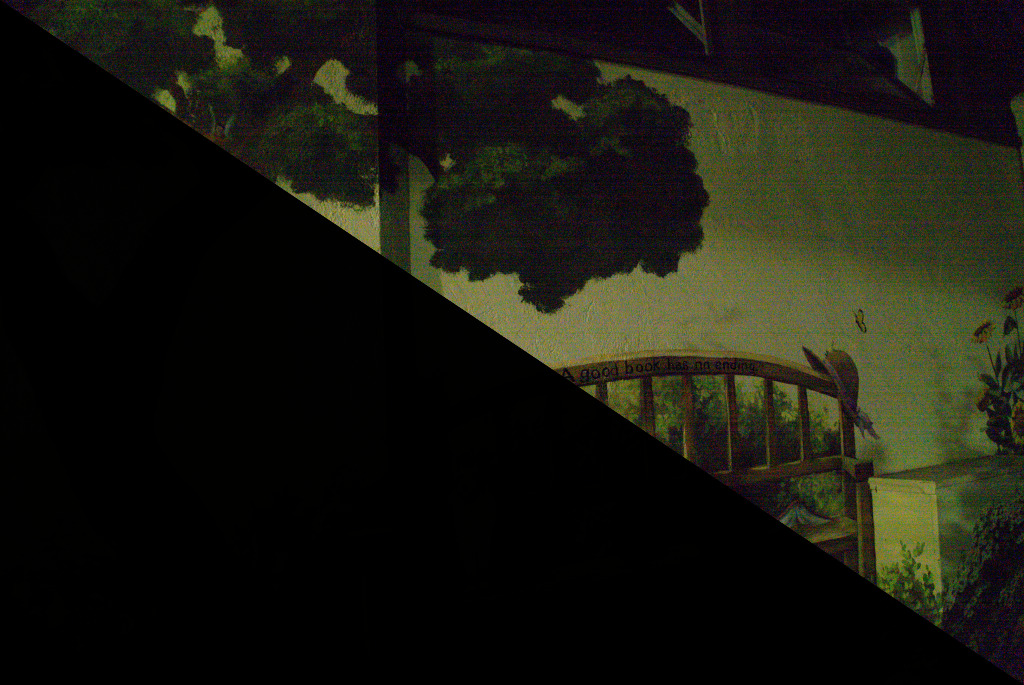}&\hspace{-4mm}
      \includegraphics[width=0.242\textwidth]{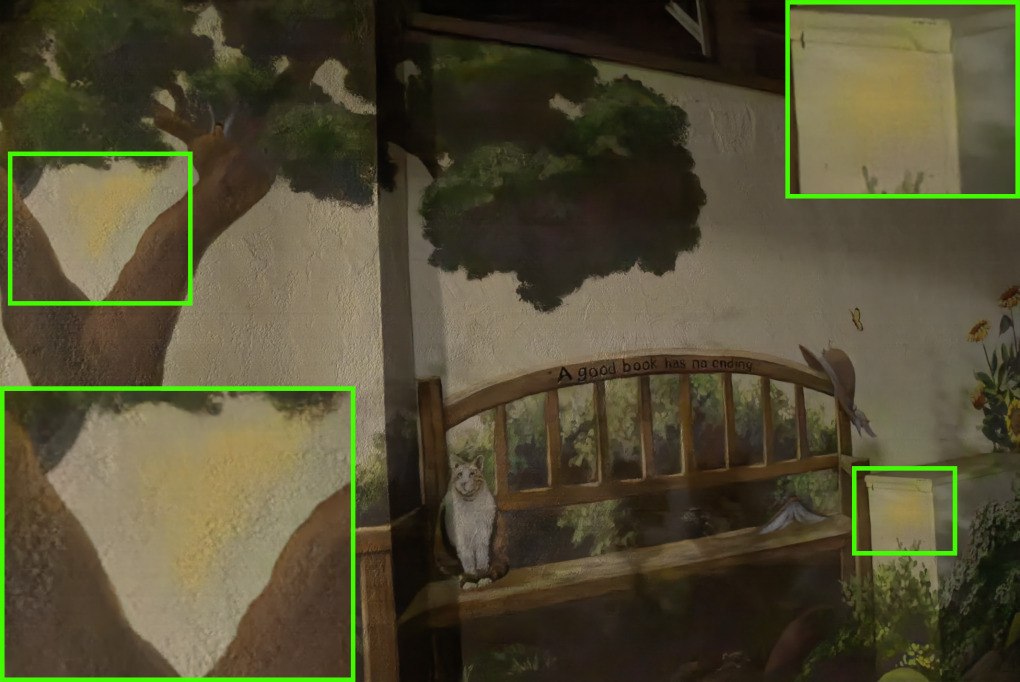}&\hspace{-4mm}
      \includegraphics[width=0.242\textwidth]{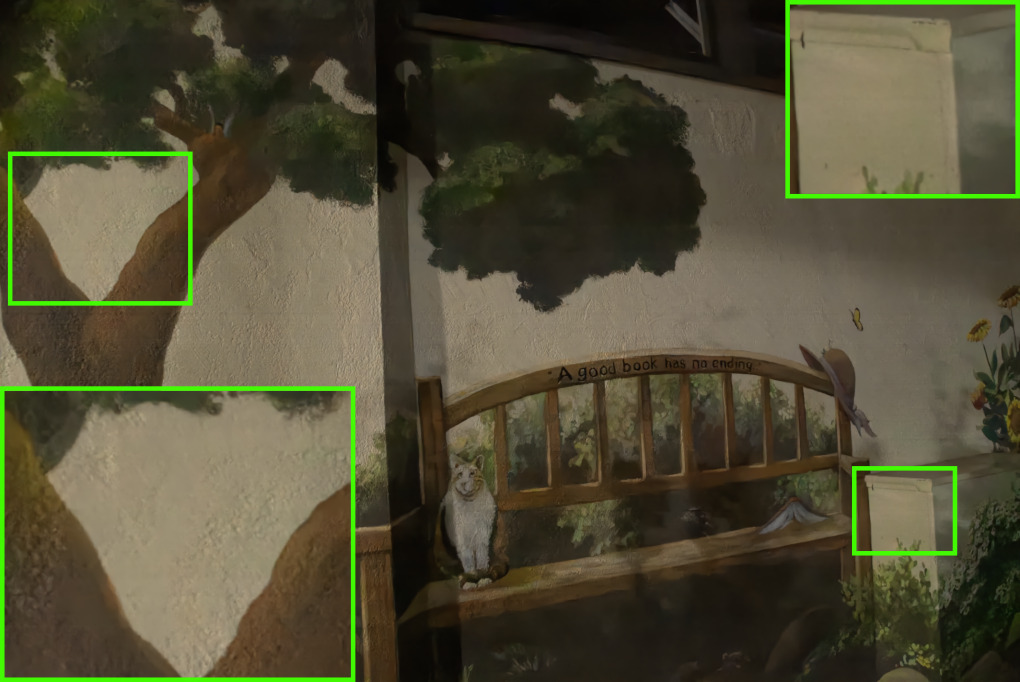}&\hspace{-4mm}
      \includegraphics[width=0.242\textwidth]{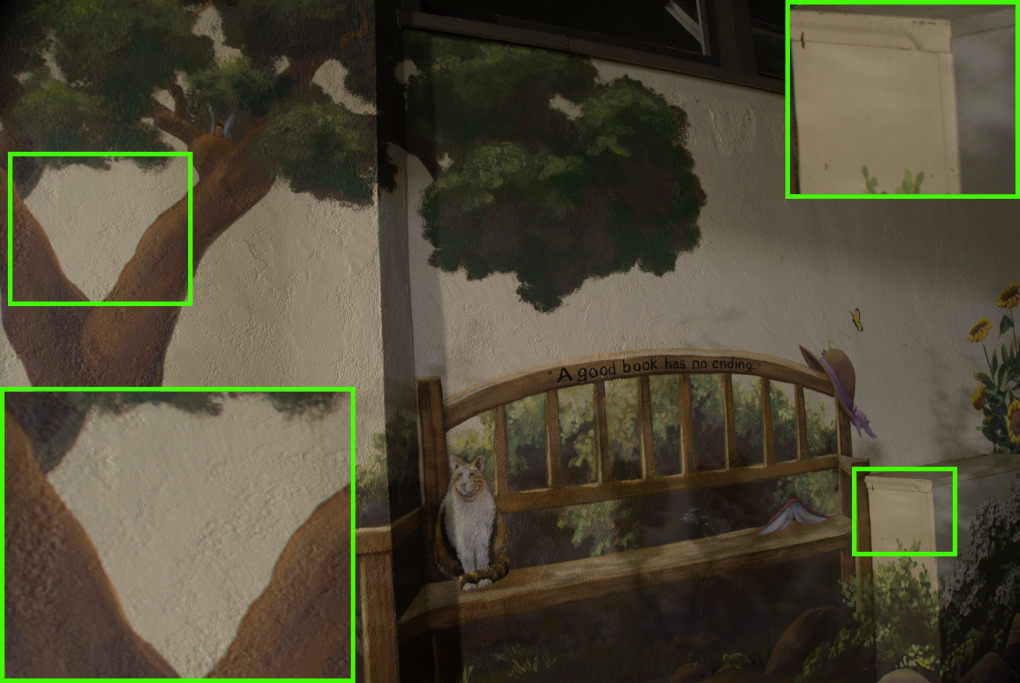}\hspace{-4mm} \\
\hspace{-6mm}
      \includegraphics[width=0.242\textwidth]{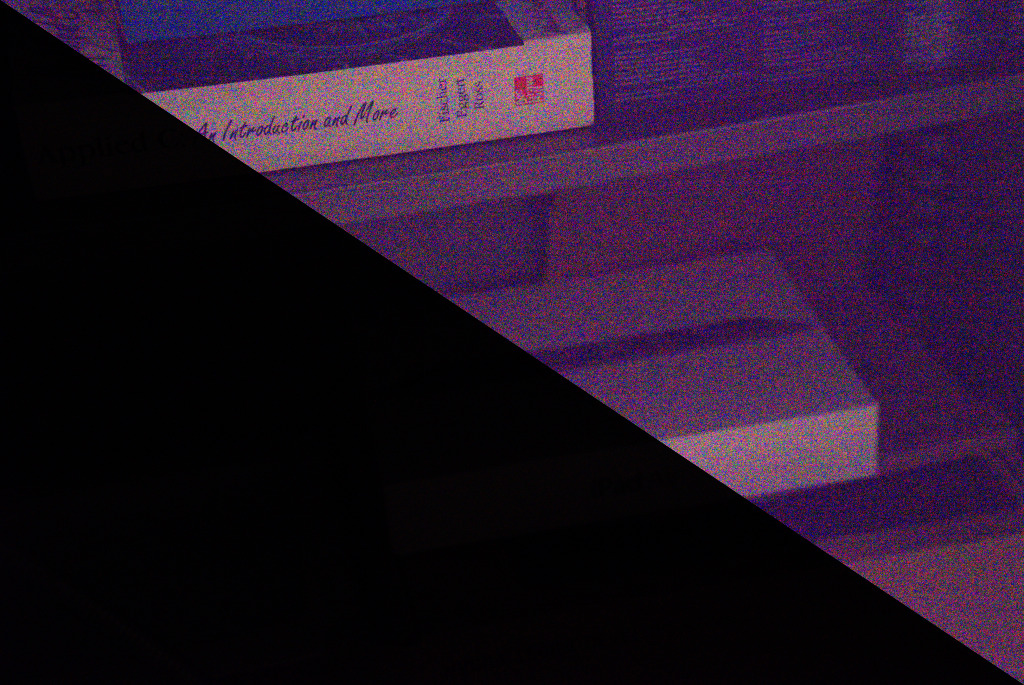}&\hspace{-4mm}
      \includegraphics[width=0.242\textwidth]{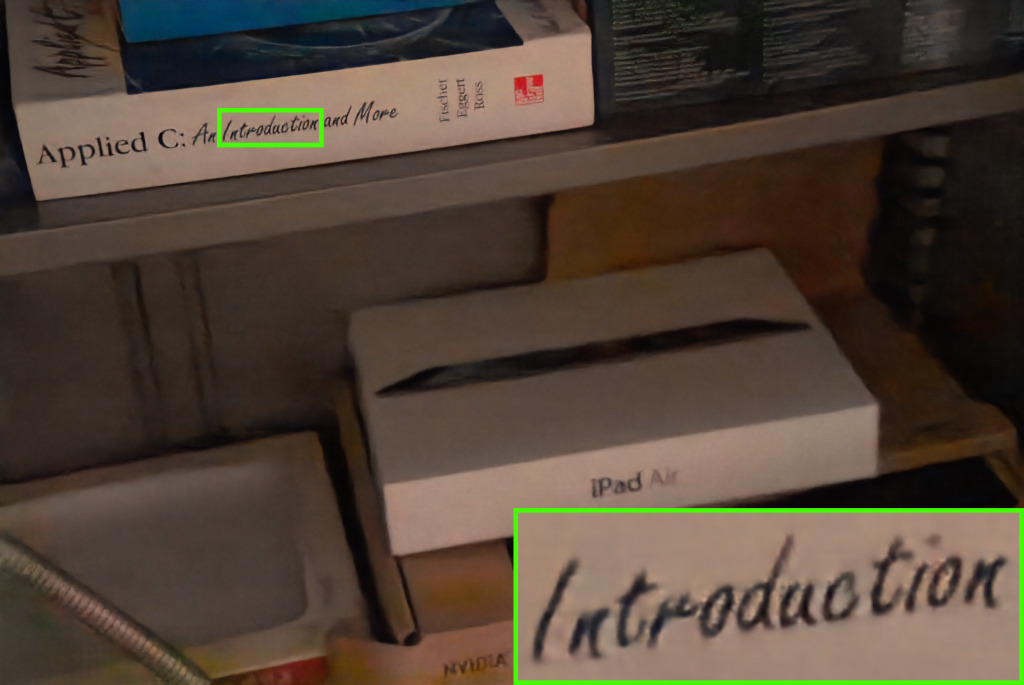}&\hspace{-4mm}
      \includegraphics[width=0.242\textwidth]{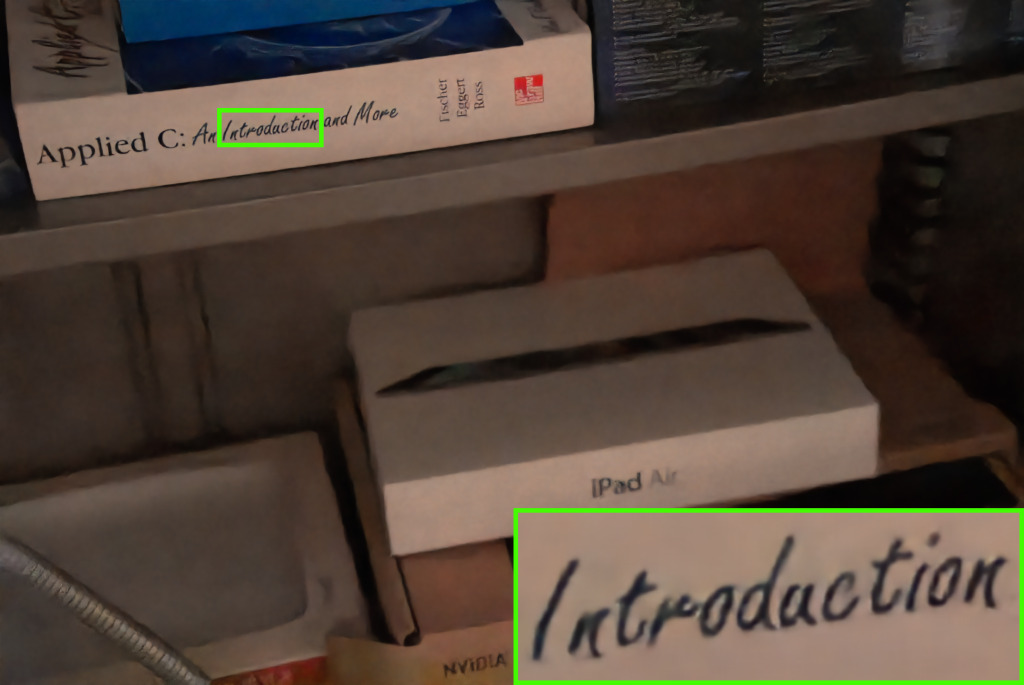}&\hspace{-4mm}
      \includegraphics[width=0.242\textwidth]{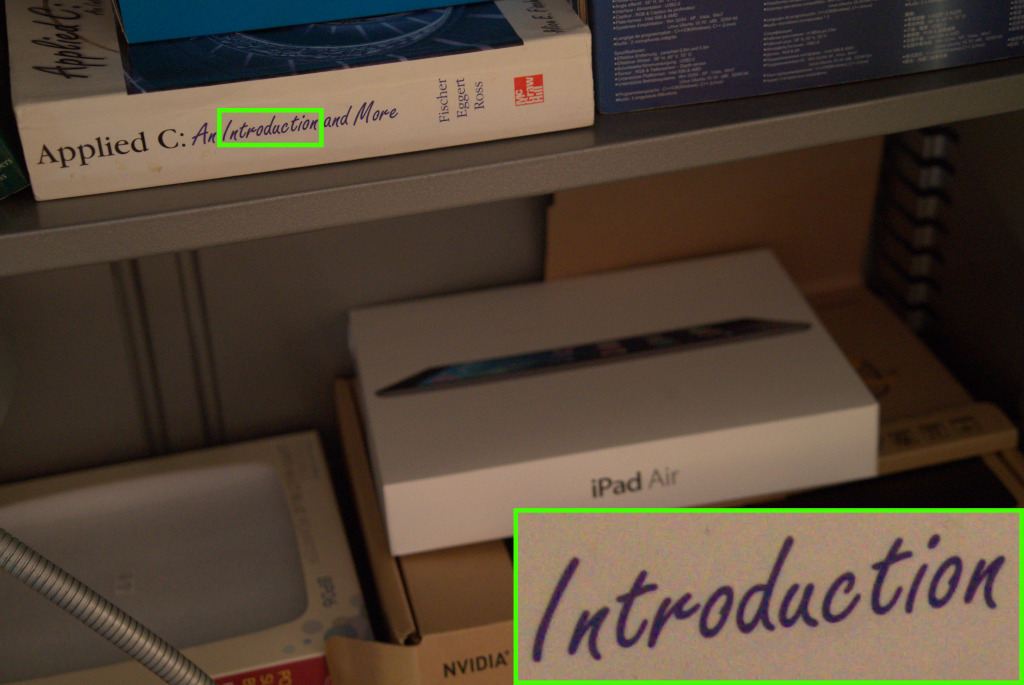}\hspace{-4mm} \\

       (a) Input & (b) Chen \etal. \cite{Chen2018}  & (c) Ours  & (d) Ground-truth \\
    \end{tabular}
  \end{center}\vspace{-2em}
      \caption{\small Visual comparisons of the proposed algorithm with the conventional camera ISP and state-of-the-art method of \cite{Chen2018} on the SID dataset.
      (a) Input image. For better visualization, the RAW image is passed through a conventional camera pipeline \cite{Ramanath2005}. 
      Note that the brightness of top right half is increased to reveal image details. 
      (b) The enhanced images by Chen \etal. \cite{Chen2018} contain artifacts (highlighted with green boxes). 
      (c) Our method is able to remove these artifacts and recover natural color and image details.
      }
    \label{Fig:qual_sid}\vspace{-1em}
\end{figure*}

\subsection{Qualitative Evaluation}
We show visual comparisons of our algorithm and the state-of-the-art on three datasets. 
Figure~\ref{Fig:qual_fivek2_2} show the images produced by our method and the state-of-the-art methods on the MIT-Adobe FiveK dataset. %
It is noticeable that our algorithm generates images that are visually more pleasant and have better local and global contrast as compared to other competing approaches.
Next, we present in Figure~\ref{Fig:qual_lol} the image enhancement results on LoL dataset. It shows that our method produces perceptually more faithful results than other algorithms. For example, compare the overall brightness, contrast and object level details.  

Finally, we show enhanced results of the proposed method under extreme low-light conditions.
Figure~\ref{Fig:qual_sid} shows the results generated by the proposed algorithm and the state-of-the-art method of Chen \etal \cite{Chen2018} on two challenging examples from the SID dataset. 
The images generated by the state-of-the-art method contain noticeable artifacts (all three zoomed-in views marked with green boxes). 
In contrast, the proposed algorithm generates images that are natural in appearance with sharper details.
We note there is no existing work that achieves consistently better results on all three datasets. It corroborates the impact of harnessing global context and local context into a single architecture that is effective for image enhancement.

\subsection{Pyschophysical Evaluation}

\begin{figure}[t]
    \centering
    \begin{subfigure}[t]{0.23\textwidth}
      \includegraphics[width=\textwidth]{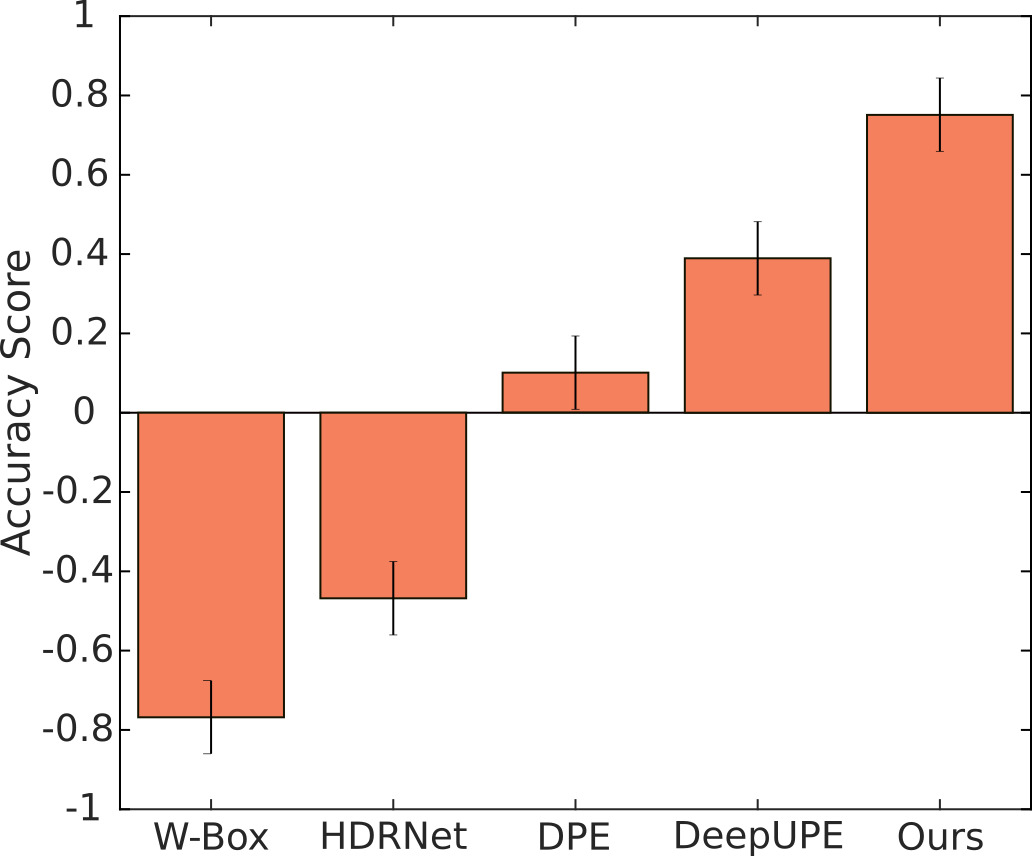}
      \caption{Results showing ground-truth to observers.}
      \label{fig:subjective accurate}
    \end{subfigure}\hspace{1mm}
    \begin{subfigure}[t]{0.23\textwidth}
      \includegraphics[width=\textwidth]{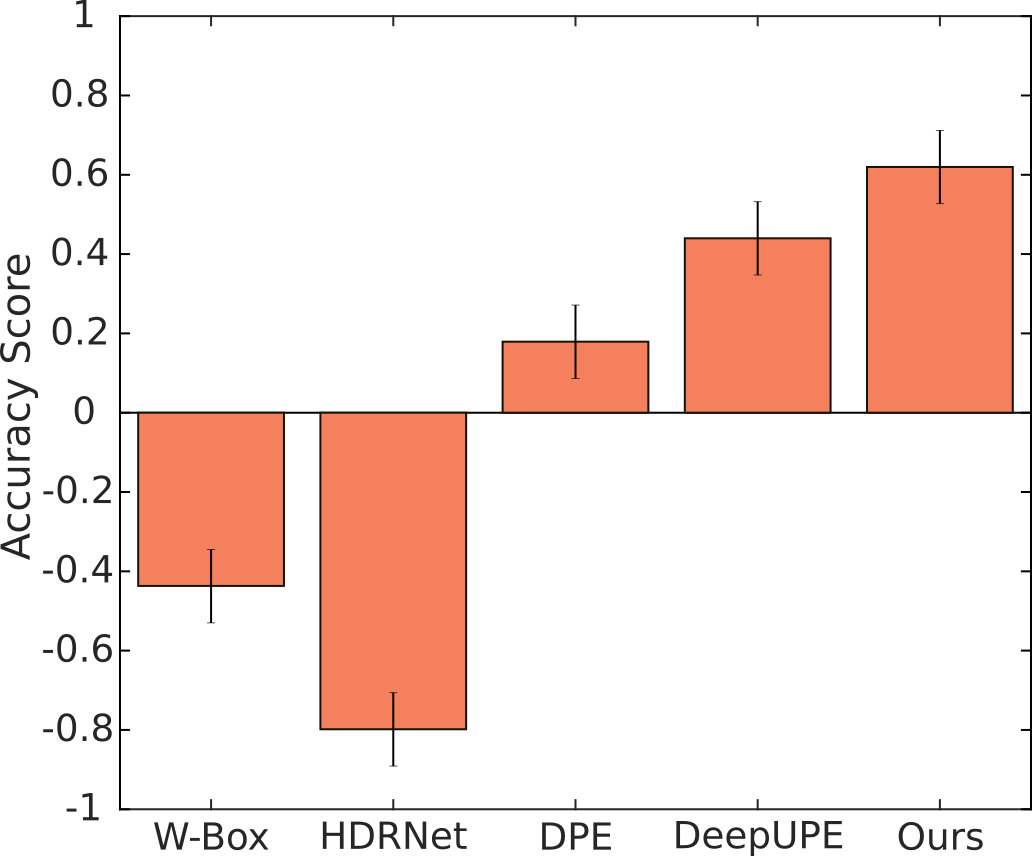}
      \caption{Results \textit{without} showing ground-truth to observers.}
      \label{fig:subjective pleasant}
    \end{subfigure}
    \vspace{-0.5em}
    \caption{\small Psychophysical Evaluation. Two experiments are performed: (a) aiming for the accurate image reproduction with respect to ground-truth, and (b) with the pleasant reproduction intent (without showing ground-truth to observers). Our method ranks first in both experiments according to the observers' choices. }
    \vspace{-1.4em}
\end{figure}

To further assess the effectiveness of our approach, we conduct user studies. 
We create a subset of 25 images by taking every 20th image from the test set of MIT-Adobe FiveK dataset. Each of these image is enhanced with our algorithm and the following four state-of-the-art approaches: W-Box~\cite{hu2018exposure}, HDRNet~\cite{Gharbi2017}, DPE~\cite{chen2018deep}, DeepUPE~\cite{wang2019underexposed}.
In the experiments, 20 observers with normal color vision participated. We used pair-wise comparison technique and each observer had to judge 250 pair of images.
For analyzing psychophysical data, we use the method of \cite{Morovic2008}, that is based on Thurstone's law of comparative judgement \cite{thurstone1927law}. 

We ran two different psychophysical experiments, where each observer was shown a pair of corresponding enhanced images (1) in isolation, and (2) alongside the ground-truth. The first psychophysical test is to investigate which algorithm produces images that the observers find most pleasant, aesthetically (color, brightness, contrast, etc.). And the second psychophysical test is to evaluate which algorithm produces images that are perceptually most faithful to the ground-truth.  
Figure~\ref{fig:subjective pleasant} shows the results for the first psychophysical experiment, where it can be seen that the observers preferred our method over other competing approaches. Furthermore, the HDRNet~\cite{Gharbi2017} ranks last in the users' preference. %
In Figure~\ref{fig:subjective accurate}, we present the accuracy scores for the second psychophysical experiment. According to the observers, our method generates images that are perceptually more closer to the ground-truth than those of the other algorithms. Moreover, it is worth noting that, for this experiment, the overall ranking trend of the competing methods is same as obtained with the image fidelity metrics.
\section{Conclusion}
We propose a context-aware hierarchical network for low-light image enhancement. The proposed approach adapts to global scene settings via modeling spatial correlations across the whole spatial extent. This is realized by introducing global context module.
We further improve context diversity by encoding relatively greater range of contextual representation in a dense manner at each resolution. 
This facilitates reliable recovery of local color and contrast while enhancing short-exposure images captured under low-light conditions. 
We perform experiments on three challenging datasets: MIT-Adobe FiveK, LoL, and SID. Results show that the proposed algorithm performs favorably against the state-of-the-art, both quantitatively and qualitatively. 
Lastly, introducing learnable context in an image enhancement network will facilitate the development of new contextual mechanisms for effective image enhancement.

\clearpage
{\small
\bibliographystyle{ieee_fullname}
\bibliography{bib}
}

\end{document}